\documentclass[letterpaper]{article}
\usepackage{graphicx}
\usepackage{pgf-pie}
\usepackage{xparse}
\usepackage{mathtools}
\usepackage{rigidnotation}
\usepackage{subfig}
\usepackage{hyperref}
\usepackage{pythonhighlight}
\NewDocumentCommand\Vector{m}{ \boldsymbol{\mathbf{#1}} }
\NewDocumentCommand\Matrix{m}{ \boldsymbol{\mathbf{#1}} }
\NewDocumentCommand\CSys{m}{ \{#1\} }

\NewRigidNotation{Vel}{v}
\NewRigidNotation{Acc}{a}

\SetConciseNotation

\title{A Standard Rigid Transformation Notation Convention for Robotics Research}

\author{Philippe Nadeau}

\usepackage{cleveref}
\begin{document}

\maketitle

\begin{abstract}
Notation conventions for rigid transformations are as diverse as they are fundamental to the field of robotics.
A well-defined convention that is practical, consistent and unambiguous is essential for the clear communication of ideas and to foster collaboration between researchers.
This work presents an analysis of conventions used in state-of-the-art robotics research, defines a new notation convention, and provides software packages to facilitate its use.
To shed some light on the current state of notation conventions in robotics research, this work presents an analysis of the ICRA 2023 proceedings, focusing on the notation conventions used for rigid transformations. 
A total of 1655 papers were inspected to identify the convention used, and key insights about trends and usage preferences are derived.
Based on this analysis, a new notation convention called \textsf{RIGID} is defined, which complies with the \textit{ISO 80000 Standard on Quantities and Units}.
The \textsf{RIGID} convention is designed to be concise yet unambiguous and easy to use.
Additionally, this work introduces a \LaTeX\ package that facilitates the use of the \textsf{RIGID} notation in manuscripts preparation through simple customizable commands that can be easily translated into variable names for software development.
\end{abstract}

\section{Introduction}

\begin{quotation}
    \textit{To form the algorithm of derivations, it was necessary to introduce new signs; I gave particular attention to this subject, convinced that the secret of the power of analysis consists in the choice and appropriate use of simple signs characteristic of the thing they are meant to represent.}
\end{quotation}
\begin{flushright}
    --- Louis François A. Arbogast in ``Du calcul des dérivations" (1800).
\end{flushright}

\subsection{Preliminaries}
In the following, a \textit{vector} is assumed to be a mathematical object that possesses both a magnitude and a direction.
A Cartesian \textit{coordinate system} is a set of three mutually orthogonal unit vectors that can be used to define the coordinates of a vector.
In robotics, it is convenient to associate \textit{reference frames} with rigid bodies to concisely describe their position and orientation, for example.

Defining a position vector requires, at the very least, specifying two elements: a \textit{subject} and a \textit{basis}.
The position vector describes the position of the subject with respect to the basis.
Additionally, specifying a coordinate system for a vector enables expressing it in terms of its coordinates, which is usually desired in robotics to facilitate computation.
Hence, an exhaustive notation for a position vector should include the subject, the basis, and the coordinate system.
Defining an orientation requires specifying two reference frames, which will be referred to as the \textit{subject} and the \textit{basis}.
With the subject reference frame fixed to a rigid body, the orientation can be described by expressing the axes of the subject with respect to those of the basis.

The role of a notation convention is to encode semantics into symbols, providing meaning to a mathematical expression.
A functional convention should allow mapping back from symbols to semantics; the encoding should be reversible.
For instance, it should be possible to interpret a given notation without ambiguity as:
\begin{itemize}
    \item the position of subject $a$ with respect to basis $b$,
    \item the position of subject $a$ with respect to basis $b$ and expressed in coordinate system $c$,
    \item the orientation of subject $a$ with respect to basis $b$,
    \item the velocity of subject $a$ relative to (or observed from) basis $b$,
    \item the force exerted on subject $a$ measured in basis $b$ and expressed in coordinate system $c$,
    \item the inertia matrix of subject $a$ computed in basis $b$,
    \item etc.
\end{itemize}

Using an exhaustive notation convention in a text should always enable the reader to understand the meaning of the symbols used, independantly of the context.
However, a minimal notation convention relies on the context to make the meaning of the symbols unambiguous.
For instance, when referring to \textit{the orientation of the object}, the author relies on the context in which the expression is used to clarify what is the basis relative to which the orientation is defined.

\subsection{Conventions Used in (a few) Robotics Books}
An idea about notation conventions used in robotics can be obtained by looking at a few robotics books.
In a subset of --- albeit arbitrarily selected --- commonly cited robotics books, the following notation conventions are used:
\begin{enumerate}
    \item $(R_{ba},p_{ba})$ in Lynch \& Park \cite{lynch_modern_2017} and also in Murray \cite{murray2017mathematical},
    \item $({}^b\boldsymbol{R}_a,{}^b\boldsymbol{p}_a)$ in Corke \cite{corke2011robotics}, Yoshikawa \cite{yoshikawa1990foundations}, and Handbook of Robotics \cite{handbook},
    \item $({}^bR^a, {}^bp^a_c)$ in Tedrake \cite{tedrake},
    \item $(\boldsymbol{C}_{ab},\boldsymbol{r}^{ba}_{c})$ in Barfoot \cite{barfoot2024state},
    \item $({}^a_b{R},{}^b{P}_a)$ in Craig \cite{craig1989introduction}.
\end{enumerate}
In the above list, the first element of the pair is the notation for the orientation of a reference frame $a$ with respect to a reference frame $b$, and the second element is the notation for the position of $a$ with respect to $b$ and expressed in a coordinate system $c$.
According to this subset of robotics books, there is no apparent consensus on the rigid transformation notation to use.
The following section presents an analysis of the ICRA 2023 proceedings that provides a more comprehensive view of the notation conventions used in the state-of-the-art of robotics research.

\section{An Analysis of ICRA 2023 Proceedings}

\subsection{Methodology}
\begin{figure}[h]
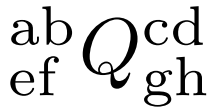

    \centering
    \scalebox{3}{%
        ${}^{\rm{ab}}_{\rm{ef}}{Q}^{\rm{cd}}_{\rm{gh}}$%
    }
    \caption{Letter code used to record the position of notational elements.}
    \label{fig:LetterCode}
\end{figure}

The archive distributed by the IEEE containing the conference proceedings consisting of 1655 PDF files was downloaded.
The title of the paper, the list of authors, and the sub-field keywords were extracted from the PDF metadata.
One by one, each PDF file was visually inspected to see if rigid transformations were used, in which case the notation convention employed was manually recorded.
The author is sincerely thankful to authors who included in their paper either a clear definition of the notation being used, a diagram showing relationships between reference frames, or a nomenclature table.
The notation was recorded according to the encoding shown in \cref{fig:LetterCode} that associates letters to positions around the physical quantity symbol.
If multiple articles had the same last author (usually assigned to the principal investigator) and also shared the same notation, only one of the papers was taken into account. 

\subsection{Statistics}
The proceedings of ICRA 2023 contain a total of 1655 articles, out of which 242 (14.6\% of all papers) make use of rigid transformations. Among these, seven papers share their notation and last author with another paper, resulting in a dataset of 235 articles for analysis.

In the dataset, a few noteworthy elements were observed. Nine papers (3.83\%) either used a notation that was difficult to decipher or employed more than one notation within the same paper, leading to inconsistencies. Additionally, three papers (1.28\%) used "C" instead of "R" as the symbol representing a rotation matrix, and six papers (2.55\%) used a left-to-right arrow in their notation.
Most papers in the dataset used a simplified notation and relied on the context to make it unambiguous. However, 11 papers (4.68\%) used an exhaustive notation for position vectors, while 154 papers (65.53\%) used an exhaustive notation for orientation.

\begin{figure}
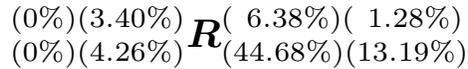

    \centering
    \scalebox{1.5}{%
        ${}^{\rm{(0\%)(3.40\%)}}_{\rm{(0\%)(4.26\%)}}\boldsymbol{R}^{\rm{(~6.38\%)(~1.28\%)}}_{\rm{(44.68\%)(13.19\%)}}$%
    }
    \caption{Proportion of papers using each location to denote the subject of an orientation.}
    \label{fig:RotLocPropA}
\end{figure}

\begin{figure}
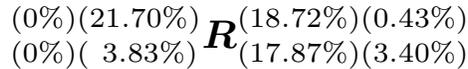

    \centering
    \scalebox{1.5}{%
        ${}^{\rm{(0\%)(21.70\%)}}_{\rm{(0\%)(~3.83\%)}}\boldsymbol{R}^{\rm{(18.72\%)(0.43\%)}}_{\rm{(17.87\%)(3.40\%)}}$%
    }
    \caption{Proportion of papers using each location to denote the basis of an orientation.}
    \label{fig:RotLocPropB}
\end{figure}

\begin{figure}
    \centering
    \subfloat[\centering Orientation]{{
        \scalebox{0.6}{%
            \begin{tikzpicture}%
                \pie[scale font]{%
                26.38/\scalebox{2}{$\boldsymbol{R}$},%
                17.45/\scalebox{2}{$\boldsymbol{R}_{a}^{b}$},%
                16.17/\scalebox{2}{${}^{b}\boldsymbol{R}_{a}$},%
                13.19/\scalebox{2}{$\boldsymbol{R}_{ba}$},%
                26.81/{\Large Others}%
                }%
            \end{tikzpicture}%
        }%
    }}
    \quad
    \subfloat[\centering Position]{{
        \scalebox{0.6}{%
            \begin{tikzpicture}%
                \pie[scale font]{%
                36.60/\scalebox{2}{$\boldsymbol{p}$},%
                16.17/\scalebox{2}{$\boldsymbol{p}_{a}^{b}$},%
                13.19/\scalebox{2}{${}^{b}\boldsymbol{p}_{a}$},%
                8.94/\scalebox{2}{$\boldsymbol{p}_{a}$},%
                25.10/{\Large Others}%
                }%
            \end{tikzpicture}%
        }%
    }}
    \caption{Most common notations to specify (a) the orientation of reference frame $a$ with respect to reference frame $b$, and (b) the position of $a$ with respect to $b$ and expressed in $c$. Note that most authors will choose to omit $c$ when $b=c$ to increase conciseness.}
    \label{fig:PieCharts}
\end{figure}
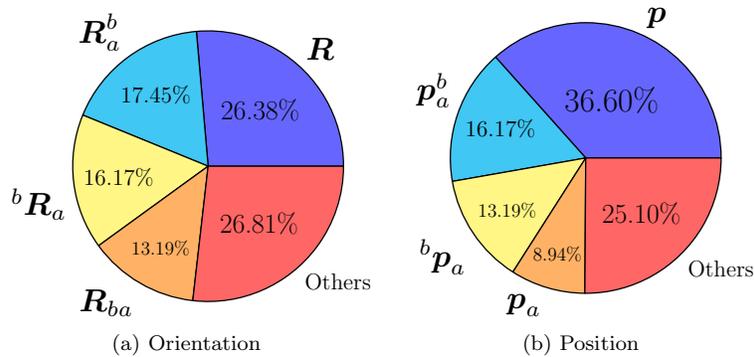

The popularity of different conventions was analyzed in two ways. 
First, the notation used was considered as a whole, and the proportion of papers using each notation was calculated, resulting in the pie charts shown in \cref{fig:PieCharts}.
Second, a finer analysis was performed to determine the proportion of papers using each location (as in \cref{fig:LetterCode}) around the quantity symbol (usually $\Matrix{R}$ or $\Vector{p}$) to specify a particular element of either the orientation or the position.
For instance, the popularity of each location when specifying the name of the frame that an orientation is defined with respect to is shown in \cref{fig:RotLocPropB} where three locations are approximately equally popular.
Note that the sum of the proportions in each figure is not necessarily 100\% as the proportions are calculated based on the total number of papers using rigid transformations, but only a subset of these papers include the information in their notation.
Hence, when analyzing \cref{fig:RotLocPropA}, \cref{fig:RotLocPropB}, \cref{fig:PosLocPropA}, \cref{fig:PosLocPropB} and \cref{fig:PosLocPropC}, the relative popularity of each location might be more informative than the absolute popularity.

To get an idea about the diversity of robotics sub-fields that make use of rigid transformations, the first keyword associated with each paper was recorded.
The complete list of papers from the dataset categorized based on their first keyword is compiled in \cref{apx:KeywordsByPopularity}, with the top 10 keywords shown in \cref{tab:Top10Keywords}.
The dominant notation used in papers with each keyword from the top 10 is also shown in \cref{tab:Top10Keywords} to give an idea of the most popular notation in each sub-field.
Note, however, that the dominant notation might not be significantly more popular than the second most popular notation.

\begin{figure}
    \centering
    \scalebox{1.5}{%
        ${}^{\rm{(0\%)(0.43\%)}}_{\rm{(0\%)(0.43\%)}}\boldsymbol{p}^{\rm{(~7.23\%)(0.43\%)}}_{\rm{(43.83\%)(6.81\%)}}$%
    }
    \caption{Proportion of papers using each location to denote the subject of a position.}
    \label{fig:PosLocPropA}
\end{figure}

\begin{figure}
    \centering
    \scalebox{1.5}{%
        ${}^{\rm{(0.43\%)(14.89\%)}}_{\rm{(0.00\%)(~2.13\%)}}\boldsymbol{p}^{\rm{(17.45\%)(1.28\%)}}_{\rm{(10.64\%)(2.55\%)}}$%
    }
    \caption{Proportion of papers using each location to denote the basis of a position.}
    \label{fig:PosLocPropB}
\end{figure}

\begin{figure}
    \centering
    \scalebox{1.5}{%
        ${}^{\rm{(0\%)(2.13\%)}}_{\rm{(0\%)(1.70\%)}}\boldsymbol{p}^{\rm{(1.70\%)(0\%)}}_{\rm{(2.55\%)(0\%)}}$%
    }
    \caption{Proportion of papers using each location to denote the coordinate system that a position vector is expressed in.}
    \label{fig:PosLocPropC}
\end{figure}

\begin{table}
    \centering
    \begin{tabular}{|l|c|c|}
        \hline
        \textbf{Keyword} & \textbf{Orientation} & \textbf{Position} \\
        \hline
        1. Localization & $\Matrix{R}_a^b$ & $\Vector{p}_a^b$ \\
        2. SLAM & $\Matrix{R}_{ab}$ & $\Vector{p}$ \\
        3. Calibration and Identification & ${}^b\Matrix{R}_a$ & ${}^b\Vector{p}_a$ \\
        4. Medical Robots and Systems & ${}^b\Matrix{R}_a$ & $\Vector{p}$ \\
        5. Sensor Fusion & ${}_a^b\Matrix{R}$ & $\Vector{p}_a^b$ \\
        6. Legged Robots & $\Matrix{R}$ & $\Vector{p}^a$ \\
        7. Perception for Grasping and Manipulation & ${}^b\Matrix{R}^a$ & $\Vector{p}$ \\
        8. Aerial Systems: Mechanics and Control & $\Matrix{R}$ & $\Vector{p}$ \\
        9. Deep Learning in Grasping and Manipulation & $\Matrix{R}$ & $\Vector{p}$ \\
        10. Visual-Inertial SLAM & ${}_b^a\Matrix{R}$ & ${}^b\Vector{p}_a$ \\
        \hline
    \end{tabular}
    \caption{Dominant notation used in papers with the top 10 keywords. Note that the dominant notation might not be significantly more popular than the second most popular notation.}
    \label{tab:Top10Keywords}
\end{table}

\subsection{Insights}
Clearly, the results in \cref{fig:PieCharts} indicate that the most popular notation is the most concise one permitted by the context --- few papers ($4.68\%$) use an exhaustive notation for position vectors.
Authors will prefer to use a minimal notation, relying on the context or on definitions provided in the text to make the meaning of the symbols unambiguous.
Sometimes, the topic of the paper will facilitate the use of a minimal notation.
For instance, when the objective is to control the position and orientation of a robot, or when the focus is on estimating the pose of a vehicle in a map.

The information in \cref{fig:PieCharts} also indicates that there is no consensus on the notation to use when the subject and basis are specified.
To specify the orientation of $a$ with respect to $b$, the notations $\Matrix{R}_a^b$, $\Matrix{R}_{ba}$, and ${}^b\Matrix{R}_a$ are all similarly popular, covering together about $50\%$ of the papers.
Amongst position notations specifying the basis, $\Vector{p}_a^b$ and ${}^b\Vector{p}_a$ are about equally popular, but cover together only about $30\%$ of the papers.

Looking closer at the popularity of each location around $\Vector{p}$ and $\Matrix{R}$ to specify the subject, it is clear from \cref{fig:RotLocPropA} and \cref{fig:PosLocPropA} that the most popular location is the right subscript ($g$ in \cref{fig:LetterCode}) with about $45\%$ of the papers using it.
The basis of the orientation is almost as often specified using the right superscript, right subscript, or left superscript, with the latter being slightly more popular.
Specifying the basis of a position vector is more often done using the right or left superscript, and to a smaller extent using the right subscript (which is commonly used to specify the subject of the position vector).
Finally, there seems to be no consensus on the location to use when specifying the coordinate system in which a position vector is expressed, as shown in \cref{fig:PosLocPropC}.

With the ten most common robotics sub-fields from the papers analyzed only covering about $40\%$ of the papers, it is clear that rigid transformations are used in a wide variety of robotics sub-fields.
Moreover, \cref{tab:Top10Keywords} shows that the dominant notation used in sub-fields varies significantly from one to another, indicating that the community does not have a consensus on the notation to use.
However, note that the dominant notation within some sub-fields seems to be semantically consistent when comparing the notations for orientation and position.
For instance, in \textit{Localization}, the notations $\Matrix{R}_a^b$ and $\Vector{p}_a^b$ are used.
In \textit{Calibration and Identification}, the notations ${}^b\Matrix{R}_a$ and ${}^b\Vector{p}_a$ are used.
To a lesser extent, in \textit{Sensor Fusion}, the notations ${}_a^b\Matrix{R}$ and $\Vector{p}_a^b$ are also somewhat semantically consistent.

From the analysis of the ICRA 2023 proceedings, it is clear that the robotics community has not reached a consensus on the notation to use for rigid transformations.
This is also supported by the diversity of notation conventions used in robotics books.
However, the community seems to prefer to minimize the number of symbols used in a notation, and will use additional symbols only when necessary to make the meaning of the expression unambiguous.
Also, the right subscript seems to be, by far, the most popular location to specify the subject of either an orientation or a position vector.
Finally, there is definitely a large consensus on the use of the letter $R$ to represent a rotation matrix, and $p$ to represent a position vector.

\section{The ISO 80000 Standard}
The International Organization for Standardization (ISO) is an independant, non-governmental effort from 171 member countries to agree on conventions and procedures to facilitate international collaboration.  
The ISO 80000-2:2019 standard on Quantities and Units (Part 2: Mathematics) \cite{ISO_80000_2} defines verbal equivalent (i.e. meaning, semantics) for commonly used mathematical symbols.
It covers a wide range of fields including combinatorics, logics, geometry, etc.
Each convention covered by the standard is numbered to make it easy to refer to.
For instance, item number 2-8.2 specifies that $a\neq b$ means ``$a$ is not equal to $b$''.
The principal elements of the standard that are relevant to rigid transformations notation conventions are summarized in \cref{tab:ISOElements} and outlined in the following.

\begin{table}[h]
    \centering
    \begin{tabular}{|c|c|c|}
        \hline
        \textbf{Element} & \textbf{Symbol} & \textbf{Meaning}\\
        \hline
        2-16.1 & $\Matrix{A}$ & Matrix with elements $a_{ij}$\\
        2-16.6 & $\Matrix{A}^{-1}$ & Inverse of $\Matrix{A}$\\
        2-16.7 & $\Matrix{A}^\mathsf{T}$ & Transpose of $\Matrix{A}$\\
        2-16.8 & $\Matrix{A}^*$ & Complex conjugate of $\Matrix{A}$\\
        2-16.9 & $\Matrix{A}^H$ & Hermitian of $\Matrix{A}$\\
        2-17.2 & $\Vector{e}_x,\Vector{e}_y,\Vector{e}_z$ & Base vectors of a Cartesian coordinate system\\
        2-9.5 & $\overrightarrow{AB}$ & Vector from $A$ to $B$\\
        2-18.1 & $\Vector{a}$, $\overrightarrow{a}$ & Vector\\
        2-18.5 & $\Vector{0}$ & Zero vector\\
        2-18.6 & $\Vector{\hat{a}}$, $\Vector{e}_{\Vector{a}}$ & Unit vector in the direction of $\Vector{a}$\\
        2-18.11 & $\Vector{a}\cdot\Vector{b}$ & Scalar product of $\Vector{a}$ and $\Vector{b}$\\
        2-18.12 & $\Vector{a}\times\Vector{b}$ & Cross-product of $\Vector{a}$ and $\Vector{b}$\\
        \hline
    \end{tabular}
    \caption{Principal elements of the ISO 80000 standard that are relevant to rigid transformations notation conventions.}
    \label{tab:ISOElements}
\end{table}

Item number 2-9.5 specifies that $\overrightarrow{AB}$ is a vector from $A$ to $B$, not to be confused with $x\rightarrow a$, which means ``$x$ tends to $a$'' according to 2-8.16.
Item number 2-9.6 specifies that $d(\rm{A},\rm{B})$ is the distance between $A$ and $B$, which corresponds to the magnitude of $\overrightarrow{AB}$. 
Note that item number 2-11.6 specifies that $C^k_n$, which is similar to one of the rotation matrix convention used in robotics, means the number of combinations without repetition of $k$ elements from a set of $n$ elements.

Item number 2-16.1 specifies that $\Matrix{A}$ is a matrix with elements $a_{ij} = \left(\Matrix{A}\right)_{ij}$, using uppercase italic boldface letters for matrices.
The inverse, transpose, complex conjugate, and Hermitian conjugate of $\Matrix{A}$ are respectively denoted by $\Matrix{A}^{-1}$, $\Matrix{A}^\mathsf{T}$, $\Matrix{A}^*$, and $\Matrix{A}^H$ according to item numbers 2-16.6, 2-16.7, 2-16.8, and 2-16.9.

Cartesian coordinates are denoted by $x,y,z$ according to item number 2-17.1 when a vector is expressed in a right-handed Cartesian coordinate system with base vectors denoted by $\Vector{e}_x,\Vector{e}_y,\Vector{e}_z$ or $\Vector{e}_1,\Vector{e}_2,\Vector{e}_3$ according to item number 2-17.2.
In the standard, a basis vector is represented as an arrow going from the origin of the coordinate system to a point one unit away in the direction of the vector.

Critically, item number 2-18.1 specifies that vectors are denoted by lowercase italic boldface letters, as in $\Vector{a}$, and that an arrow above the letter can be used instead of the boldface, as in $\overrightarrow{a}$.
The zero vector, whose magnitude is zero, is denoted by $\Vector{0}$ or $\overrightarrow{0}$ according to item number 2-18.5.
A unit vector in the direction of $\Vector{a}$ is denoted by $\Vector{\hat{a}}=\Vector{e}_{\Vector{a}}=\Vector{a}/\vert\Vector{a}\vert$ by item number 2-18.6.
The usual scalar product of vectors $\Vector{a}$ and $\Vector{b}$ is denoted by $\Vector{a}\cdot\Vector{b}$ and the cross-product by $\Vector{a}\times\Vector{b}$ according to item numbers 2-18.11 and 2-18.12 respectively. 

\section{The Definition of a RIGID Notation}
The \textsf{RIGID} notation defined in this document is aimed at being (1) compliant with ISO 80000, (2) well-defined, (3) unambiguous, (4) concise, and (5) easy to use.
To be compliant with the ISO 80000 standard, the constraints imposed by the standard are considered in the first place by eliminating possibilities of conventions that would not allow to comply with the standard.
This document is aimed at being normative, such that referencing this document in a paper is sufficient to dissipate any confusion a reader might have about the notation used in the paper.
A concise form of the notation is defined alongside rules for its use, such that the meaning of the symbols used in the notation is unambiguous even when a shortened form is used.
To favor ease of use in papers and softwares, a \LaTeX\ package with simple commands used to typeset the notation is provided alongside a convention on how the \textsf{RIGID} notation should be used when defining variable names in software source code.
Additionally, an open-source reference frame management tool that facilitate defining poses and computing rigid transformations is made available.

When using this convention in your paper, the following paragraph can be included to succinctly define the notation:
\begin{quotation}
    In the following, the orientation of $\CSys{s}$ with respect to $\CSys{b}$ is denoted by $\Rot{s}{b}$, and the position of $\CSys{s}$ with respect to $\CSys{b}$ and expressed in $\CSys{c}$ is denoted by $\Pos{s}{b}{c}$, as defined by the RIGID notation convention [\textit{link to this document}].
\end{quotation}

For a notation convention to be compliant with the ISO 80000 standard, it should allow
\begin{figure*}[h!]
\begin{enumerate}
    \item common operations on vectors and matrices (e.g. inverse, transpose) to occupy the right superscript position,
    \item a hat symbol above the quantity symbol to denote a unit vector,
    \item a left-to-right arrow above te quantity symbol to denote a vector when boldface is not available, and
    \item the right subscript to denote the endpoint of a unit vector.
\end{enumerate}
\label{fig:ISOConstraints}
\end{figure*}

The left-to-right arrow prescribed by the ISO 80000 standard implies that the origin of the vector is on the left while the endpoint is on the right.
This is consistent with the endpoint of a unit vector being denoted by the right subscript in the standard.
Hence, it makes sense to use the right subscript to denote the subject of a vector and the left subscript to denote its basis, such that the arrow points from the basis to the subject.
Although the standard prescribes that the arrow be placed above the quantity symbol, the same space might be occupied by other symbols, such as the hat symbol for unit vectors or a dot for time derivatives.
Hence, this convention permits the use of the left-to-right arrow under the quantity symbol when boldface cannot possibly be used (a slight deviation from the standard that should not be commonplace since boldface is widely available in modern typesetting systems).
The left superscript can be used to denote the coordinate system in which a vector is expressed, as this location is not occupied by any elements of the standard.
The resulting notation convention, in its exhaustive form, is shown in \cref{fig:NotationDefinition}.

\begin{figure}[h]
    \centering
    \subfloat[\centering Vectors]{{
        \scalebox{3}{%
            $\Pos{s}{b}{c} \text{ , } {}^{c}_{b}\underrightarrow{p}_{s}$
        }
    }}
    \quad\quad
    \subfloat[\centering Orientation]{{
        \scalebox{3}{%
            $\Rot{s}{b} \text{ , } {}_{b}\underrightarrow{R}_{s}$
        }
    }}
    \caption{Notation convention for vectors with $s$ being the subject, $b$ being the basis, and $c$ being the coordinate system. When boldface is not available, a left-to-right arrow under the quantity symbol can be used instead.}
    \label{fig:NotationDefinition}
\end{figure}

Note that swapping the position of the basis and coordinate system in the notation could lead to confusion when the  arrow is used since the origin of the arrow would be closer to the coordinate system than to the basis.
Also, note that the position of the basis symbol in the notation for the orientation could be moved to the left superscript position, which is unoccupied.
This would be done at the expense of clarity since the location of the basis would differ between the notation for the orientation and the position.
Indeed, with this convention when the basis corresponds to the coordinate system, ${}_{b}\boldsymbol{Q}_{s}$ can be interpreted as \textit{$Q$ of $s$ with respect to $b$} independantly of the whether $Q$ refers to an orientation or a position vector.

In many situations, a concise form of the notation can be used without creating any ambiguity.
The concise forms of the notation, along with the conditions for their use, are defined in \cref{tab:ExhaustiveConciseNotation}.
For example, one could write
\begin{align*}
    \Rot{e}{f} \Pos{a}{e} &= \Pos{a}{e}{f}\\
    \Pos{f}{e} + \Pos{g}{f}{e} &= \Pos{g}{e}\\
    \Rot{f}{e}\Rot{g}{f}\Rot{h}{g} &= \Rot{h}{e}
\end{align*}
without any ambiguity when using this convention as the conditions for the use of the concise notation can be assumed to be met.

\begin{table}
    \centering
    \begin{tabular}{|c|c|p{0.6\textwidth}|}
        \hline
        \textbf{Exhaustive} & \textbf{Concise} & \textbf{Condition for Concise Use}\\
        \hline
        \multicolumn{3}{|c|}{\textbf{The orientation of subject $s$ with respect to basis $b$}}\\
        $\Rot{s}{b}$ & & None\\
        $\Rot{s}{b}$ & $\Rot{s}$ & There is a single basis $b$.\\
        $\Rot{s}{b}$ & $\Rot$ & There is a single subject $s$ and a single basis $b$.\\
        \multicolumn{3}{|c|}{ \textbf{The position of subject $s$ with respect to basis $b$}}\\
        \multicolumn{3}{|c|}{ \textbf{and expressed in coordinate system $c$.}}\\
        $\Pos{s}{b}{c}$ & & None\\
        $\Pos{s}{b}{c}$ & $\Pos{s}{b}$ & $b=c$\\
        $\Pos{s}{b}{c}$ & $\Pos{s}$ & There is a single basis $b=c$.\\
        $\Pos{s}{b}{c}$ & $\Pos$ & There is a single subject $s$ and a single coordinate system $b=c$.\\ 
        \hline
    \end{tabular}
    \caption{Exhaustive and concise notation for the orientation and position of a subject $s$ with respect to a basis $b$ and expressed in a coordinate system $c$, with the conditions for the use of the concise notation.}
    \label{tab:ExhaustiveConciseNotation}
\end{table}

A quantity that is derived from another one should keep the same level of information in its notation.
For instance, the \textit{pose} of a rigid body describes both its position and orientation.
Therefore, the notation for the pose must combine elements from the notation of the position and orientation.
Following the rules in \cref{tab:ExhaustiveConciseNotation}, the pose of a reference frame $s$ fixed on a rigid body with respect to a reference frame $b$ and whose position is expressed in a coordinate system $c$ would be denoted by
\begin{equation*}
    \Pose{s}{b}{c} = \begin{bmatrix}
        \Rot{s}{b} & \Pos{s}{b}{c}\\
        \Vector{0} & 1
    \end{bmatrix}~,
\end{equation*}
which forms a $4\times 4$ matrix.
However, according to \cref{tab:ExhaustiveConciseNotation}, when $b=c$, the notation can be simplified to
\begin{equation*}
    \Pose{s}{b} = \begin{bmatrix}
        \Rot{s}{b} & \Pos{s}{b}\\
        \Vector{0} & 1
    \end{bmatrix}~,
\end{equation*}
in which the notation of the pose still contains the same level of information as the notation of the position and orientation.

Another example of a derived quantity is the linear momentum of a point mass, which is the product of its mass and velocity.
The linear momentum should be denoted with the same amount of information as the velocity.

\subsection{Usage in \LaTeX\ Files}
The \textsf{rigidnotation} \LaTeX\ package was developed to make it easy to typeset vectors and matrices following the RIGID convention. 
To do so, the package defines a \textit{command factory} that can be used to concisely generate multiple commands, each one dedicated to a user-defined quantity (e.g. position vector, velocity vector, rotation matrix). 
The user can then call the generated commands within math-mode environments to typeset a concise representation of the quantity.
For instance, \verb|\Pos[\dot]{s}{b}{w}\Tran| produces
\begin{equation*} 
 \resizebox{1cm}{!}{$\Pos[\dot]{s}{b}{w}\Tran$}~,
\end{equation*}
which very concisely represents the time-derivative of the position vector of subject $s$ with respect to basis $b$ when expressed in coordinate system $w$. 
When typing the keystrokes, the user can think: \textit{position of $s$ with respect to $b$ expressed in $w$, transposed}. 
The order in which the elements are typed respects the order in which they can be spoken, making it easier to remember the order of the arguments.

Additionally, a \textit{concise mode} can be enabled such that when the basis and coordinate system are the same, only the basis is denoted.

The typical usage of the package involves defining quantities of interest in the \LaTeX\ document preamble, like 
\begin{verbatim}
  \usepackage{rigidnotation}
  \NewRigidNotation{Vel}{v} % Velocity
  \NewRigidNotation{Acc}{a} % Acceleration
  % Whether the concise notation is desired
  \SetConciseNotation{\BooleanTrue}
\end{verbatim}
in which quantities representing velocities and accelerations are defined.
While the first line imports the package, the second and third lines create new commands (\verb|\Vel| and \verb|\Acc| respectively) that can be called with up to four arguments, and whose quantity symbols are $\boldsymbol{v}$ and $\boldsymbol{a}$ respectively.
Then, in any math-mode environment, variations of the following commands can be used:
\begin{verbatim}
  \begin{equation*}
    1:\Vel \quad 
    2:\Vel[\hat] \quad
    3:\Vel[\ddot]{s} \quad 
    4:\SetConciseNotation \Vel{s}{b}{b} \quad
    5:\Vel{s}{b}{c} \quad
    6:\UnsetConciseNotation \Vel{s}{b}{b} \quad
    7:\Acc{s}{b}\Tran \quad 
    8:\Acc{s}{b}^2 \quad
    9:\Rot{s}{b}\Inv \quad 
  \end{equation*}
\end{verbatim}
to produce
\begin{equation*}
    1:\Vel \quad 
    2:\Vel[\hat] \quad
    3:\Vel[\ddot]{s} \quad 
    4:\SetConciseNotation \Vel{s}{b}{b} \quad
    5:\Vel{s}{b}{c} \quad
    6:\UnsetConciseNotation \Vel{s}{b}{b} \quad
    7:\Acc{s}{b}\Tran \quad 
    8:\Acc{s}{b}^2 \quad
    9:\Rot{s}{b}\Inv \quad
  \end{equation*}
where the concise notation was enabled for items 4 and 5 but disabled for element 6 (notice the difference). 
Also note that command \verb|\Rot| was used even though it was not defined in the preamble. 
This is because the package comes with a few pre-defined quantities: \verb|\Pos|, \verb|\Rot|, and \verb|\Pose|.

The \textsf{rigidnotation} package is available on CTAN at:
\begin{quote}\vspace{-2mm}\centering\url{https://ctan.org/pkg/rigidnotation}\vspace{-2mm}\end{quote}
and development files are open-sourced on GitHub at
\begin{quote}\vspace{-2mm}\centering\url{https://github.com/PhilNad/rigidnotation-latex},\vspace{-2mm}\end{quote}
any contributions are welcome.
More documentation on the package can be found in the \textsf{rigidnotation} manual available on CTAN.

\subsection{Usage in Source Code}
Using mathematical notation in source code is a common issue caused by the fact that most programming languages only allow naming variables with alphanumeric characters and underscores.
This is especially problematic in robotics research where the notation can be complex and verbose.

In an effort to tackle this issue, the team behind the Drake simulator developed \href{https://drake.mit.edu/doxygen_cxx/group__multibody__quantities.html}{a naming convention} (coined \textit{monogram}) that adapts their notation convention such that it can be used in source code.
Also, the authors of the Orocos Kinematics and Dynamics Library \href{http://docs.ros.org/en/indigo/api/orocos_kdl/html/geomprim.html}{also tackled this issue in their software} and converged to a similar convention.
In both cases, the first character of the variable name is the quantity symbol and underscore characters are used to separate the elements of the notation.

The RIGID convention can be used within source code in a similar way.
However, in contrast with the \textit{monogram} notation used in Drake, the order of the elements in the variable name respects the order in which they can be spoken with \textit{quantity of $s$ with respect to $b$ expressed in $c$}, which also corresponds to the order in which they are typed with the \textsf{rigidnotation} \LaTeX\ package.
For instance, the position of subject $s$ with respect to basis $b$ and expressed in coordinate system $c$ could be denoted by \texttt{p\_s\_b\_c}.
Another slight variation from the \textit{monogram} notation is the use of an underscore between the subject and basis.
As a result, a \LaTeX\ command easily translates to a variable name in source code, as shown in \cref{tab:LaTeXToSourceCode}, and allows to directly reuse the rules for concise notation outlined in \cref{tab:ExhaustiveConciseNotation}.

\begin{table}[h]
    \centering
    \begin{tabular}{|c|c|}
        \hline
        \textbf{Typesetting Code} & \textbf{Programming Code}\\
        \hline
        \verb|\Pos{s}{b}{c}| & \texttt{p\_s\_b\_c}\\
        \verb|\Pos[dot]{s}{b}| & \texttt{pdot\_s\_b}\\
        \verb|\Rot{s}{b}\Inv| & \texttt{R\_s\_b\_Inv}\\
        \verb|\v[dot]{s}{b}{c}\Tran| & \texttt{vdot\_s\_b\_c\_Tran}\\
        \hline
    \end{tabular}
    \caption{Translation of \LaTeX\ commands to variable names in source code.}
    \label{tab:LaTeXToSourceCode}
\end{table}

Notational elements that are placed above the quantity symbol in the typeset notation are placed directly after the quantity symbol in the variable name, with no underscore directly after the quantity.
To further make the variable name be as close as possible to the \LaTeX\ command, a notational element at the exponent position is appended to the variable name with an underscore separating it from the rest of the variable name using the abbreviations in \cref{tab:ExponentAbbreviations}.

\begin{table}[h]
    \centering
    \begin{tabular}{|c|c|}
        \hline
        \textbf{Notational Element} & \textbf{Abbreviation}\\
        \hline
        Transpose & Tran \\
        Inverse & Inv \\
        Complex Conjugate & Conj \\
        Hermitian & Herm \\
        \hline
    \end{tabular}
    \caption{Abbreviations for notational elements at the exponent position.}
    \label{tab:ExponentAbbreviations}
\end{table}

With this scheme, it should be possible to easily produce \textsf{rigidnotation} commands from variable names, and vice versa.
This comes at the cost of disallowing the use of the abbreviations in \cref{tab:ExponentAbbreviations} as names for the subject, basis, or coordinate system.

\subsubsection{A Library to Stop worrying about Rigid Transformations}
Computing rigid transformations between various reference frames is a common task that is prone to errors due to the variety of representations and conventions.
Although technically simple, the task of memorizing poses and computing transformations can be time-consuming and hinder research progress.
The \href{http://wiki.ros.org/tf2}{\textsf{tf2} ROS package} can be used to manage reference frames for a robotic system, but requires a ROS environment to be used, which can be cumbersome if the user is not working on a ROS project.

To alleviate the issues related to managing reference frames potentially shared by multiple software components, the \href{https://github.com/PhilNad/with-respect-to}{\textsf{With-Respect-To}} library was developed.
The library is designed to be simple to use, fast to execute, and accessible from a variety of interfaces.
The simplicity of the library is achieved by essentially providing a total of two features: (1) setting a pose and (2) getting a pose.
The library is designed to be very fast such that the user does not (usually) have to worry about the computational cost of using it.
Stress tests on basic hardware have shown that the library can manage references frames at a rate of 300-400 Hz on very large kinematic chains and with multiple concurrent readers and writers.
Finally, the library can be used with Python, C++, and through a command-line interface, making it easy to integrate into a variety of projects --- an example of how the library can be used in Python is shown in \cref{lst:WRTExample}.
Since the library maintains a database of reference frames, it can be used to manage reference frames across multiple concurrent processes with efficiency.
The \textsf{With-Respect-To} library is open-source and available on GitHub at \url{https://github.com/PhilNad/with-respect-to}.

\begin{lstlisting}[style=mypython,caption={Usage of the library in Python},label=lst:WRTExample]
import numpy as np
import with_respect_to as WRT

#Connect/create a reference frame database
db = WRT.DbConnector()

#Rotation of 90 deg. around the x-axis
X_b_a = np.array([[1,0,0,0],[0,0,-1,0],[0,1,0,0],[0,0,0,1]])
db.In('test').Set('b').Wrt('a').Ei('a').As(pose)

#Translation of [1,1,1]
X_a_w = np.array([[1,0,0,1],[0,1,0,1],[0,0,1,1],[0,0,0,1]])
db.In('test').Set('a').Wrt('w').Ei('w').As(pose)

#Rotation of 90 deg. around the z-axis and translation of [1,1,0]
X_c_b = np.array([[0,-1,0,1],[1,0,0,1],[0,0,1,0],[0,0,0,1]])
db.In('test').Set('c').Wrt('b').Ei('b').As(pose)

#Get the pose of C with respect to W expressed in A
X_c_w_a = db.In('test').Get('c').Wrt('w').Ei('a')

\end{lstlisting}

\appendix

\section{Keywords By Popularity}
\label{apx:KeywordsByPopularity}
This list enumerates the most popular keywords for papers that made use of rigid transformations in the ICRA 2023 proceedings.
Note that since multiple keywords were allowed for each paper, only the first keyword was taken into account.
The percentage in parentheses indicates the proportion of papers, amongst those that used rigid transformations, that selected the keyword as their first keyword.
With the top 10 keywords covering about 40\% of the 242 papers, clearly rigid transformations are used in a wide variety of robotics sub-fields.
\begin{enumerate}
    \item \textbf{Localization} (7.7\%): \cite{65, 181, 299, 394, 701, 790, 947, 1203, 1270, 1414, 1814, 1861, 2974, 3088, 3560, 3692, 3693, 3700}
    \item \textbf{SLAM} (5.1\%): \cite{6, 114, 161, 188, 1471, 2436, 2757, 3171, 3188, 3206, 3615, 3665}
    \item \textbf{Calibration and Identification} (4.7\%): \cite{200, 378, 437, 445, 614, 1053, 1054, 1126, 1356, 2173, 3670}
    \item \textbf{Medical Robots and Systems} (4.7\%): \cite{607, 624, 834, 838, 1165, 1166, 1167, 1557, 2737, 3237, 3350}
    \item \textbf{Sensor Fusion} (3.8\%): \cite{48, 64, 918, 1213, 1523, 2474, 2660, 3197, 3497}
    \item \textbf{Legged Robots} (3.8\%): \cite{1449, 1763, 2073, 2565, 2817, 3065, 3151, 3161, 3289}
    \item \textbf{Perception for Grasping and Manipulation} (3.4\%): \cite{95, 133, 1443, 1519, 2547, 2758, 3061, 3557}
    \item \textbf{Aerial Systems: Mechanics and Control} (3.0\%): \cite{110, 263, 1441, 1715, 1743, 2263, 3329}
    \item \textbf{Deep Learning in Grasping and Manipulation} (2.6\%): \cite{73, 85, 362, 2116, 2344, 3039}
    \item \textbf{Visual-Inertial SLAM} (2.6\%): \cite{120, 952, 1316, 1605, 3616, 3643}
    \item \textbf{Marine Robotics} (2.6\%): \cite{588, 820, 981, 2107, 2744, 3658}
    \item \textbf{Modeling, Control, and Learning for Soft Robots} (2.6\%): \cite{671, 1260, 1598, 2811, 3516, 3655}
    \item \textbf{Human-Robot Collaboration} (2.1\%): \cite{132, 748, 1564, 2159, 3075}
    \item \textbf{Vision-Based Navigation} (2.1\%): \cite{176, 398, 1371, 1675, 2060}
    \item \textbf{Deep Learning for Visual Perception} (2.1\%): \cite{178, 366, 565, 2341, 2608}
    \item \textbf{Object Detection, Segmentation and Categorization} (2.1\%): \cite{213, 418, 1190, 1926, 2844}
    \item \textbf{Aerial Systems: Applications} (1.7\%): \cite{102, 912, 1586, 3710}
    \item \textbf{Industrial Robots} (1.7\%): \cite{105, 2597, 2694, 3133}
    \item \textbf{Rehabilitation Robotics} (1.7\%): \cite{160, 1527, 1925, 3479}
    \item \textbf{Computer Vision for Automation} (1.7\%): \cite{162, 2824, 3022, 3098}
    \item \textbf{Visual Servoing} (1.7\%): \cite{656, 1860, 2016, 2031}
    \item \textbf{Dexterous Manipulation} (1.7\%): \cite{711, 2203, 2533, 3142}
    \item \textbf{Machine Learning for Robot Control} (1.7\%): \cite{1515, 1789, 1869, 1915}
    \item \textbf{Micro/Nano Robots} (1.3\%): \cite{60, 269, 854}
    \item \textbf{Mapping} (1.3\%): \cite{184, 492, 3617}
    \item \textbf{Biologically-Inspired Robots} (1.3\%): \cite{660, 2993, 3231}
    \item \textbf{Surgical Robotics: Steerable Catheters/Needles} (1.3\%): \cite{1182, 3021, 3103}
    \item \textbf{Field Robots} (1.3\%): \cite{2013, 2042, 2755}
    \item \textbf{Space Robotics and Automation} (1.3\%): \cite{2819, 2834, 2876}
    \item \textbf{Physical Human-Robot Interaction} (0.9\%): \cite{130, 2091}
    \item \textbf{Force and Tactile Sensing} (0.9\%): \cite{175, 1973}
    \item \textbf{Redundant Robots} (0.9\%): \cite{456, 885}
    \item \textbf{Kinematics} (0.9\%): \cite{839, 3649}
    \item \textbf{Constrained Motion Planning} (0.9\%): \cite{925, 1393}
    \item \textbf{Parallel Robots} (0.9\%): \cite{1138, 3481}
    \item \textbf{Motion and Path Planning} (0.9\%): \cite{1155, 1736}
    \item \textbf{Cellular and Modular Robots} (0.9\%): \cite{1305, 2700}
    \item \textbf{Aerial Systems: Perception and Autonomy} (0.9\%): \cite{2149, 2914}
    \item \textbf{Representation Learning} (0.9\%): \cite{2678, 3236}
    \item \textbf{Simulation and Animation} (0.9\%): \cite{2973, 3644}
    \item \textbf{Mechanism Design} (0.9\%): \cite{3632, 3661}
    \item \textbf{Learning from Demonstration}: \cite{74}
    \item \textbf{Cognitive Modeling}: \cite{84}
    \item \textbf{Data Sets for SLAM}: \cite{121}
    \item \textbf{RGB-D Perception}: \cite{150}
    \item \textbf{Robust/Adaptive Control}: \cite{151}
    \item \textbf{Humanoid Robot Systems}: \cite{183}
    \item \textbf{Humanoid and Bipedal Locomotion}: \cite{218}
    \item \textbf{Manipulation Planning}: \cite{280}
    \item \textbf{Semantic Scene Understanding}: \cite{436}
    \item \textbf{In-Hand Manipulation}: \cite{525}
    \item \textbf{Mobile Manipulation}: \cite{530}
    \item \textbf{Data Sets for Robotic Vision}: \cite{576}
    \item \textbf{Compliant Joints and Mechanisms}: \cite{644}
    \item \textbf{Swarm Robotics}: \cite{645}
    \item \textbf{Optimization and Optimal Control}: \cite{741}
    \item \textbf{Robotics and Automation in Agriculture and Forestry}: \cite{765}
    \item \textbf{Methods and Tools for Robot System Design}: \cite{856}
    \item \textbf{Tendon/Wire Mechanism}: \cite{876}
    \item \textbf{Visual Tracking}: \cite{1061}
    \item \textbf{Reinforcement Learning}: \cite{1150}
    \item \textbf{Planning under Uncertainty}: \cite{1206}
    \item \textbf{Integrated Planning and Control}: \cite{1253}
    \item \textbf{Climbing Robots}: \cite{1895}
    \item \textbf{Prosthetics and Exoskeletons}: \cite{2011}
    \item \textbf{Telerobotics and Teleoperation}: \cite{2072}
    \item \textbf{AI-Based Methods}: \cite{2115}
    \item \textbf{Omnidirectional Vision}: \cite{2133}
    \item \textbf{Multifingered Hands}: \cite{2451}
    \item \textbf{Learning Categories and Concepts}: \cite{2718}
    \item \textbf{Task and Motion Planning}: \cite{2764}
    \item \textbf{Autonomous Vehicle Navigation}: \cite{2767}
    \item \textbf{Multi-Robot SLAM}: \cite{2924}
    \item \textbf{Grasping}: \cite{3086}
    \item \textbf{Collision Avoidance}: \cite{3125}
    \item \textbf{Computer Vision for Medical Robotics}: \cite{3176}
    \item \textbf{Swarms}: \cite{3341}
    \item \textbf{Origami robot}: \cite{3345}
    \item \textbf{Multi-Contact Whole-Body Motion Planning and Control}: \cite{3521}
    \item \textbf{Compliance and Impedance Control}: \cite{3588}
    \item \textbf{Haptics and Haptic Interfaces}: \cite{3633}
    \item \textbf{Robotics and Automation in Life Sciences}: \cite{3646}
    \item \textbf{Flexible Robots}: \cite{3662}
    \item \textbf{Visual-Based Navigation}: \cite{3699}
\end{enumerate}

\bibliographystyle{plain}
\bibliography{icra2023.bib}

\end{document}